\documentclass[conference,10pt]{IEEEtran}
\usepackage{geometry}
\usepackage{algpseudocode}
\algtext*{EndWhile}
\algtext*{EndIf}
\algtext*{EndFor}
\algtext*{EndProcedure}
\usepackage{amsmath}
\usepackage{amssymb}
\usepackage{array}
\usepackage{algorithm}
\usepackage{bm}
\usepackage[bookmarks=false]{hyperref}
\usepackage[noadjust]{cite}
\usepackage[font=small]{caption}
\usepackage{color}
\usepackage{comment}
\usepackage{dblfloatfix}
\usepackage[inline]{enumitem}
\usepackage{epsfig}
\usepackage{etoolbox}
\usepackage{fancybox}
\usepackage{float}
\usepackage{fixltx2e}
\usepackage{graphicx}
\usepackage{hyperref}
\usepackage{listings}
\usepackage{mathrsfs}
\usepackage{multirow}
\usepackage{multicol}
\usepackage{pifont}
\usepackage{placeins}
\usepackage{rotating}
\usepackage{setspace}
\usepackage{soul}
\usepackage[hang, tight]{subfigure}
\usepackage{threeparttable}
\usepackage{times}
\usepackage{url}
\usepackage{upgreek}
\usepackage{verbatim}
\usepackage{wrapfig}
\usepackage[usenames,dvipsnames]{xcolor}
\usepackage{authblk}
\usepackage[]{siunitx}
\usepackage[super]{nth}
\usepackage{tabularx}
\usepackage{supertabular}
\usepackage{tikz}
\usepackage{bbm}
\usepackage{blindtext}

\newcommand{\RUDY}{\mbox{$\mathit{RUDY}$}}

\newcommand{\OP}{\mbox{$\mathit{OP}$}}
\newcommand{\Concat}{\mbox{$\mathit{Concat}$}}

\newtheorem{problem}{Problem}

\graphicspath{{./_fig/}}

\begin{document}


\title{Automatic Routability Predictor Development \\ Using Neural Architecture Search \vspace{-2mm}}


    

\author[1]{\normalsize Chen-Chia~Chang\textsuperscript{*}}
\author[1]{Jingyu~Pan\textsuperscript{*}}
\author[1]{Tunhou~Zhang}
\author[1]{Zhiyao~Xie}
\author[2]{Jiang~Hu}
\author[3]{Weiyi~Qi}
\author[4]{ \\ Chun-Wei~Lin} 
\author[2]{Rongjian~Liang}
\author[3]{Joydeep~Mitra}
\author[3]{Elias~Fallon}
\author[1]{Yiran~Chen} 

\affil[$1$]{\normalsize Duke~University, Durham, NC, USA}
\affil[$2$]{Texas A\&M University, College Station, TX, USA}
\affil[$3$]{Cadence, TX, USA; $^4$National Taiwan University, Taipei, TW}
\affil[$1$]{\{chenchia.chang, jingyu.pan, tunhou.zhang, zhiyao.xie\}@duke.edu}

\maketitle
\begingroup\renewcommand\thefootnote{*}
\footnotetext{These two authors contributed equally to this manuscript.}
\endgroup

\begin{abstract}

The rise of machine learning technology inspires a boom of its applications in electronic design automation (EDA) and helps improve the degree of automation in chip designs. 
However, manually crafted machine learning models require extensive human expertise and tremendous engineering efforts.
In this work, we leverage neural architecture search (NAS) to automate the development of high-quality neural architectures for routability prediction, which can help to guide cell placement toward routable solutions. Our search method supports various operations and highly flexible connections, leading to architectures significantly different from all previous human-crafted models.
Experimental results on a large dataset demonstrate that our automatically generated neural architectures clearly outperform multiple representative manually crafted solutions. 
Compared to the best case of manually crafted models, NAS-generated models achieve 5.85\% higher Kendall's $\tau$ in predicting the number of nets with DRC violations and 2.12\% better area under ROC curve (ROC-AUC) in DRC hotspot detection. Moreover, compared with human-crafted models, which easily take weeks to develop, our efficient NAS approach finishes the whole automatic search process with only 0.3 days.

\end{abstract}

\section{Introduction}

Modern digital IC design is usually a large engineering project that consists of many interacting complicated steps. 
Although EDA tools have highly automated most design steps, many significant imperfections and challenges persist in existing design methodologies. For example, an early-step solution might turn out to work poorly in practice, as it lacks a credible prediction for its impact on subsequent design steps. 
As a result, designers need to spend many design iterations to reach an optimized design quality, and this largely increases the overall turnaround time. 


Machine learning (ML) techniques have been popularly adopted to improve the interactions between design steps by enabling early-stage predictions~\cite{ChanISPD17,TabriziDAC18,xie2018routenet,YuDAC19,LiangISPD20,chen2020pros}.
For example, ML models are applied to predict whether decisions at early design steps will lead to satisfactory design objectives in subsequent steps. 
With fast feedback from these ML models, a design converges to a high-quality solution with significantly fewer iterations than traditional EDA flows, and the overall turnaround time is shortened. 
In existing works, convolutional neural network (CNN) models~\cite{xie2018routenet, yu2019pin,LiangISPD20} and generative adversarial network (GAN) models~\cite{YuDAC19, alawieh2020high} are popular choices of the applied ML models. 
However, the development of models is challenging as it requires extensive expertise and tremendous engineering efforts on both ML and EDA.
For example, designing a neural network architecture for industrial applications can easily take months to complete for an experienced developer. 
Such challenge significantly prolongs the development cycle of ML-based EDA tools and greatly exacerbates EDA development/application cost.

Automated machine learning (AutoML), especially neural architecture search (NAS), enables design automation of a large variety of ML models without (or with minimum) human interventions.
Neural network architectures provided by NAS~\cite{pham2018efficient} have outperformed state-of-the-art manual designs with significantly improved model accuracy and computation efficiency.
Given a target task such as image classification and segmentation, NAS firstly identifies an architecture search space and then applies certain search strategies, such as reinforcement-learning-based~\cite{zoph2018learning} methods or evolutionary-guided~\cite{lu2019nsga} methods to judiciously discover promising architectures.
As chip layouts can be represented and processed like images~\cite{xie2018routenet, YuDAC19}, it is natural to use CNN-based models for layout applications and leverage associated NAS techniques to automate the ML model development process.



In this work, we propose a NAS-based method to automate the development of ML models for routability prediction.
Routability prediction estimates the routability of design solutions at the placement stage~\cite{li2007routability} with the following two application scenarios: \textit{violated net count prediction}~\cite{xie2018routenet} and \textit{DRC hotspot detection}~\cite{xie2018routenet, chen2020pros, YuDAC19}. 
\textit{Violated net count prediction} evaluates the number of nets with DRC violations in the entire layout.
\textit{DRC hotspot detection} identifies the locations of design rule violations, which can be used to guide DRC violation mitigation techniques.
Our NAS search space is abstracted as graphs that allow rich and flexible interactions among components within the model to better capture congestion patterns.
Then we obtain the best model in the graph search space by adopting sub-graph sampling as our search strategy.



Our main contributions are summarized as follows:

\begin{itemize}
\item We propose a NAS-based method to automatically develop routability estimators without human interference. 
To the best of our knowledge, our NAS-based methodology is the first research effort on automatic ML development for EDA problems. 
\item We design a large search space allowing various types of operations and highly flexible connections, some of which are never adopted in previous routability estimators. This ensures diversity in its candidate models.
\item Our NAS-crafted model outperforms several representative routablity estimators~\cite{xie2018routenet, chen2020pros, YuDAC19} by 5.8\% in Kendall's $\tau$ and 9.7\% in correlation for violated net count prediction, and achieves 2.1\% better ROC-AUC for DRC hotspot detection.
This is evaluated on a comprehensive dataset, which comprises more than 7,000 layouts from 74 designs.
\item Our NAS method is highly efficient in automatic model development. The whole search process takes only 0.3 days while the human developers easily spend weeks to months for a promising model.
\item We provide a detailed analysis on the search output, i.e. the NAS-crafted model, which differs significantly from human-developed estimators. This may benefit future routablity estimator development. 
\end{itemize}


\section{Preliminaries}
\subsection{Machine Learning for Routability Prediction} \label{sec:pre_CNN ResNet}

Early routability prediction enables designers or EDA tools to perform preventive measures such that DRC violations can be avoided in a proactive manner. 
This is a representative topic in ML for EDA, with its benefit to chip quality well demonstrated in many previous works~\cite{xie2018routenet, yu2019pin,LiangISPD20,chen2020pros, qi2014accurate, chan2016beol}. 
In recent years, deep neural network methods including CNN and fully convolutional network (FCN) become the dominant solutions~\cite{xie2018routenet, yu2019pin,LiangISPD20,chen2020pros} to routability prediction. 
The CNN models are typically used for violation number prediction~\cite{xie2018routenet}. 
As for DRC hotspot detection, which requires pinpointing specific locations with DRC hotspots in a two-dimensional layout, it shares a similar setting with semantic segmentation in identifying pixel-wise properties. 
Thus, the FCN, as a popular technique for semantic segmentation, is used in DRC hotspot detection~\cite{xie2018routenet, chen2020pros, LiangISPD20}.

Among representative routability estimators in recent years, RouteNet~\cite{xie2018routenet} and J-Net~\cite{LiangISPD20} propose U-Net-like FCN structures, PROS~\cite{chen2020pros} adopts an encoder-decoder FCN framework, and Yu et al.~\cite{YuDAC19} propose a conditional generative adversarial network~\cite{YuDAC19} (cGAN)-based model. 
To the best of our knowledge, all previous routability estimators ~\cite{xie2018routenet, yu2019pin,LiangISPD20,chen2020pros, qi2014accurate, chan2016beol} are designed by human developers. Thus, they require both ML and EDA expertise and easily take weeks of model development time.
In addition, previous works develop their models mostly based on hierarchical structures, with a limited number of branch structures. 
In comparison, our graph-based search space enables highly flexible connections and rich branches, thus providing significantly different model structures. 
The details of our search space are presented in Section~\ref{sec:NAS}. 
Since the complex pattern behind routability prediction may be reflected by complicated interactions among features in a wide layout region, branch structures can capture combined information from different sources and benefit model performance. 

%

\subsection{Neural Architecture Search}
\begin{figure}[t]
\centering
\includegraphics[width=\linewidth]{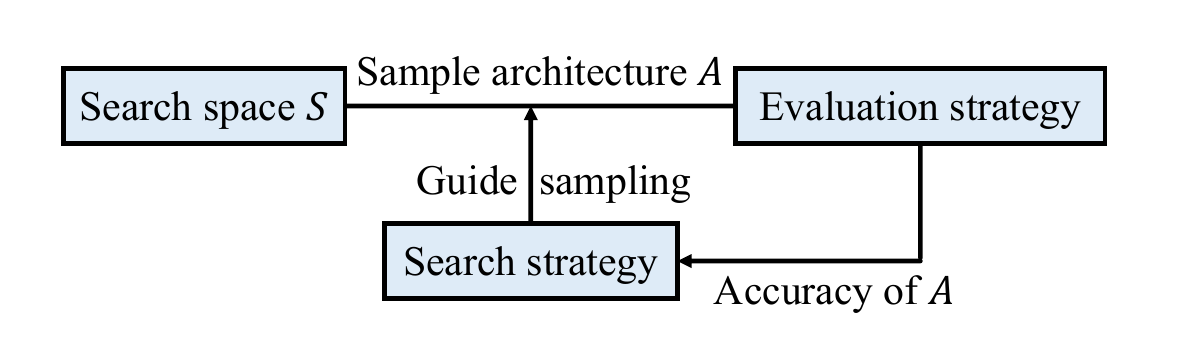}
\caption{Overview of Neural Architecture Search. }
\label{fig:overview of NAS}
\vspace{-12pt}
\end{figure}

Neural architecture search~\cite{zoph2018learning} automatically conducts architecture engineering to find effective and efficient neural network models for specific tasks without (or with minimum) human interventions.
Recent works demonstrate great potential of NAS in applications including image classification~\cite{zoph2018learning}, object detection~\cite{zoph2018learning}, and semantic segmentation~\cite{liu2019auto}.
NAS contains three key ingredients: \textit{search space}, \textit{evaluation strategy}, and \textit{search strategy}.
Search space defines a family of candidate architectures that can be explored in NAS. Evaluation strategy determines the way to estimate the design metrics (e.g., accuracy) of a candidate architecture and provides feedback to the search process. Search strategy is the method to explore the search space and guide the search process toward a correct choice of the promising ML model. 
The overall procedure of NAS is sketched in Fig.~\ref{fig:overview of NAS}.


Our NAS approach abstracts neural architecture search space into graphs and includes various types of operations as search options. 
Then, our NAS approach performs proxyless evaluation~\cite{cai2018proxylessnas} to evaluate the performance of candidate models on the target dataset. Finally, our NAS approach develops a search strategy based on a progressively graph updating and sampling algorithm~\cite{cheng2019swiftnet} to efficiently explore ML models.

\begin{figure*} [!t]
\centering
\includegraphics[width = \textwidth]{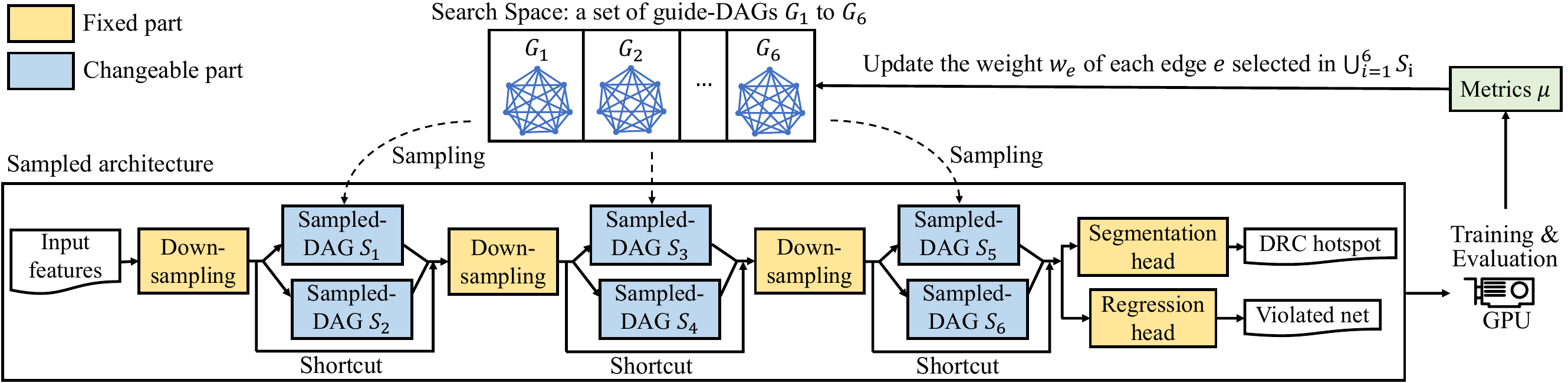}
\caption{
Overview of our graph-based NAS method. We sample multiple DAGs from the search space to form a sampled model. After training for several epochs, we get the evaluation metrics, i.e., Kendall ranking coefficient or ROC-AUC, and utilize it to update the weight of each edge in guide-DAGs.
}
\vspace{-1em}
\label{figure:nas ea flow}
\end{figure*}

\section{Problem Formulation}

We apply NAS techniques to assist the design of ML models for routability prediction. 
After placement, a layout is tessellated into $w\times h$ tiles, then its input feature $X_i \in \mathbb{R}^{w\times h \times c}$ is comprised of $c$ different two-dimensional feature maps. 
The ground-truth label is collected after detailed routing finishes. Based on the extracted features and label, the routability prediction tasks can be formulated below:

\begin{problem}
[\textbf{NAS for violated net count prediction}]
Given a set of placement solutions and the defined search space $S_C$, it aims to generate the input feature $X_i$, explore the architecture $A_C \in S_C$ of the neural network model $f_{A_C}$
to predict the violated net count $y_i$ such that the performance of $f_{A_C}$ is maximized, where
$$
f_{A_C} : X_i\in \mathbb{R}^{w\times h \times c} \rightarrow y_i
\in \mathbb{R}.
$$
\label{pro:violated net number}
\end{problem}


\begin{problem}[\textbf{NAS for DRC hotspot detection}]
Given a set of placement solutions, it aims to generate the input feature $X_i$ and the defined search space $S_L$, explore the architecture $A_L \in S_L$ of the neural network model $f_{A_L}$ 
to detect the locations of DRC hotspots $Y_i$ such that the performance of $f_{A_L}$ is maximized, where
$$
f_{A_L} : X_i\in \mathbb{R}^{w\times h\times c} \rightarrow Y_i
\in \{ 0, 1 \}^{w \times h}.
$$
\label{pro:DRC hotspot prediction}
\end{problem}


The performance of model $f_{A_C}$ is evaluated by Kendall ranking coefficient $\tau$, while the performance of $f_{A_L}$ is evaluated by ROC-AUC~\cite{hanley1982meaning}. 
The Kendall's $\tau \in [-1, 1]$ captures the rank-based correlation between the violated net count labels and predictions from $f_{A_C}$.
A higher $\tau$ indicates that $f_{A_C}$ ranks layouts more accurately (identical if $\tau=1$).
Receiver operating characteristic (ROC) curve plots the tradeoff between true positive rate versus false positive rate by varying classification threshold.
We use the area under the ROC (ROC-AUC) as the metrics of diagnostic ability of the model.
A higher ROC-AUC indicates that higher precision of DRC hotspot detection can be achieved at the cost of the same number of false alarms.


\section{Methodology} \label{sec:algorithm}

In this section, we first introduce our neural architecture search procedure, including search spaces, evaluation strategies, and search strategies for both violated net count prediction and DRC hotspot detection. Then, we present the input features extracted from placement solutions for routability prediction. 



\subsection{Neural Architecture Search} \label{sec:NAS}
We propose a graph-based NAS method motivated by~\cite{cheng2019swiftnet} to automate the design of neural networks for high-quality violated net count prediction and DRC hotspot detection. In the following, we introduce three key components of our NAS method in detail: 1) search space, 2) evaluation strategy, 3) search strategy.

\noindent \textbf{Search Space.} 

In our NAS-based model, we can partition the architecture into two parts: one part is iteratively changed during the search process, while the other is fixed. 
Our model is shown in Fig.~\ref{figure:nas ea flow}.
The yellow rectangles represent the fixed part with widely-adopted structures, and the six blue rectangles indicate the changeable part. 
In the following, we demonstrate the architectures of the fixed part and the changeable part.

In the fixed part, as Fig.~\ref{figure:nas ea flow} shows, there are three downsampling layers, which are convolution layers with a stride of 2 to reduce feature maps. 
For violated net count prediction, there is one fixed regression head at the end. 
This regression head is a mean pooling layer followed by two dense layers with output sizes 32 and 1, which maps the output from the previous blocks to a scalar representing predicted violated net count. 
For DRC hotspot detection, the structure at the end is the segmentation head. 
It is composed of three transposed convolution layers to recover feature maps with a total upsampling factor of 8 and one convolution layer to compress feature maps to a DRC hotspot solution. 
This is a two-dimensional output that pinpoints locations with DRC hotspots.
In addition, we add one shortcut between every two downsampling layers to further boost the performance as proposed in the famous ResNet~\cite{he2016deep} model.

As for the changeable part, we start with deciding all candidate operations in the search space. 
First, regular convolution layers with different numbers of filters are included.
Besides, atrous convolution, also named dilated convolution~\cite{chen2017rethinking}, is selected as a promising candidate operation since it can effectively enlarge receptive fields of filters. 
This operation can thus help to capture large patterns, such as congestions caused by nets spanning a large region. 
In addition, the work of~\cite{tan2019mixconv} introduces a new mixed depth-wise convolution (MixConv) that separates channels into groups and applies different kernel sizes to each group.
Compared with a regular convolution that can only observe patterns in a fixed size area, this operation can identify congestion patterns of different sizes when applied in routability prediction. 
Thus, MixConv is a good fit for our work since routability can be affected by the relations of nets and standard cells in different regions within a layout. 
Among these candidate operations, to our best knowledge, atrous convolution is only adopted in a recent routability estimator~\cite{chen2020pros}, and MixConv is never used in routability predictions. 
Adopting various promising operations can improve diversity in candidate models and help cover more potential high-quality models in the search space. 
As a result, the candidate operations $\OP$ include the following four types:
    \begin{itemize}
        \item $3 \times 3$ convolution with $32$ filters
        \item $3 \times 3$ convolution with $64$ filters
        \item $3 \times 3$ atrous convolution with dilation rate $2$, $32$ filters
        \item mixed convolution with $4$ groups, kernel size $[7, 9, 11, 13]$
    \end{itemize}

\begin{figure*} [t]
    \centering
\includegraphics[width = 0.9\textwidth]{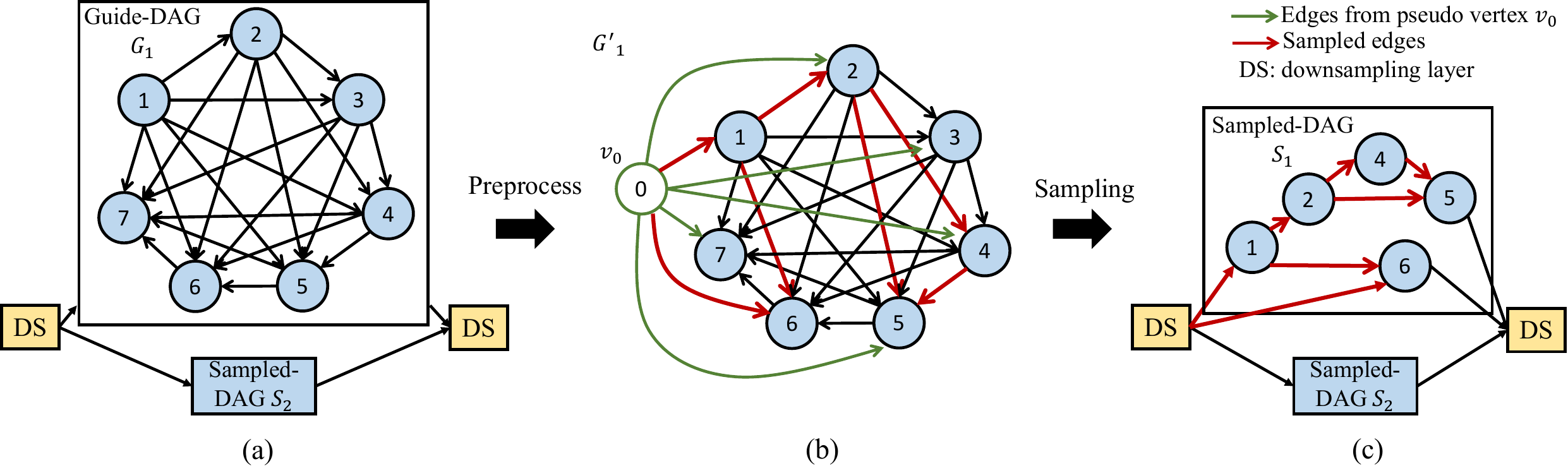}
\caption{
An example of the subgraph sampling. (a) A guide-DAG $G_1$. (b) $G'_1$ after preprocessing.
(c) Sampled-DAG $S_1$.
}
\label{figure:subgraph example}
\end{figure*}

By viewing CNN/FCN as a set of operations and the connections of operations, a model can be regarded as a graph. 
Specifically, vertices represent operations and edges indicate the directed connections of operations.
Therefore, we view the six blue changeable parts $\{S_1$, $S_2$, \dots , $S_6\}$ in Fig.~\ref{figure:nas ea flow} as DAGs, named sampled-DAGs.
There are two parallel sampled-DAGs between every two downsampling layers. 
The changes of them are restricted and guided by six guide-DAGs  $\{G_1$, $G_2$, \dots, $G_6\}$. 
Each guide-DAG $G_i(V_i, E_i)$ represents a combination of the candidate operations and the propagation of data tensors. 
It is composed by a set of completely ordered vertices $V_i$, with each vertex $v \in V_i$ representing a candidate operation $\OP_v$. 
Each edge $e(u, v) \in E_i$ represents the propagation of the output tensor of vertex $u$ to the input of $v$.
Edge $e(u, v)$ is constructed if $u < v$ in their order, which makes the guide-DAG $G_i$ completely ordered with maximum edges to provide all possible connections.
Fig.~\ref{figure:subgraph example} (a) shows an example of a guide-DAG, where the complete order of vertices is $1\rightarrow 2 \rightarrow \dots \rightarrow 7$.
Each vertex concatenates all its input tensors from the incoming edges and produces the output tensor by its operation.
Specifically, given a vertex $v$ with $\OP_v$ and input tensors $i_{u_1}, i_{u_2}, \dots, i_{u_k}$ from incoming edges $e(u_1, v), e(u_2, v), \dots, e(u_k, v)$, the output tensor $o_v$ of this vertex is 
$$
o_v = \OP_v(\Concat(i_{u_1}, i_{u_2}, \dots, i_{u_k})).
$$
Concatenation with all the input tensors of each vertex can help the model to discover different feature combinations to enhance the observation of routability information.



In our search space, the parallel sampled-DAG structures between every two downsampling layers can produce different feature representations, and pass their aggregation to the next sampled-DAG structures after downsampling.
More importantly, our search space can develop many parallel propagations of tensors within a sampled-DAG since the topology of the graph $G_i$ contains all possible connections.
In summary, our method provides a large search space and sufficient flexibility to seek for more different features representations and enable high performance routability prediction.

\vspace{0.4em}\noindent \textbf{Evaluation Strategy.} 

Previous work~\cite{cai2018proxylessnas} suggests to conduct NAS on the target dataset to improve the performance of our searched models.
Thus, we directly apply search on the training split of our target dataset and use the performance on the validation split as the search objective.
More specifically, for violated net count prediction, we use the Kendall's $\tau$ evaluated on the validation split as the search objective.
For DRC hotspot detection, we use ROC-AUC~\cite{hanley1982meaning} as the search objective.

\vspace{0.4em} \noindent \textbf{Search Strategy.} 
\begin{algorithm}[!t]
 \caption{Connection weights updating in the meta graph}
 \label{alg:weight_update}
\begin{algorithmic}[1]
\Require A set of guide-DAGs $\{G_i, i=1$ to $6\}$, baseline metrics $\beta$, learning rate $\alpha$
\For{$i=1$ to $6$}
    \State $G'_i = $ Preprocess$(G_i)$  \Comment{pseudo vertex $v_0$ insertion}
\EndFor
 \While {$\eta$ not converge}
 \For{$i=1$ to $6$}
          \State $S_i = $ Sampling$(G'_i)$ \Comment{Algorithm.~\ref{alg:sample_dag}}
      \EndFor
    \State $M = $ ConstructModel$( \{ S_i, i=1$ to $6 \} )$
    \State $\eta = $ Eval$(M)$ 
    \For{$i=1$ to $6$}
    \For{all edge $e$ in $G'_i$} 
        \If {$e$ is selected in $S_i$}
                \State $ w_e = w_e * \exp{ (\alpha (\eta - \beta))}$
        \Else
                \State $ w_e = w_e$
        \EndIf
    \EndFor
    \EndFor
    \State $\beta = $ average of top five metrics of all sampled graphs
\EndWhile 
\end{algorithmic} 
\end{algorithm}

Given the large size of our search space, exhaustively examining every subgraph in the search space is neither efficient nor practical.
Our solution is to sample edges from the guide-DAG $G_i$ with probabilities and define a weight on each edge to control its sampling probability.
We will gradually update weights through our search process to find a promising model within the search space.
The flow of our search strategy is detailed in Algorithm~\ref{alg:weight_update}.
First, in the preprocessing step (line~1-2), 
we construct a pseudo vertex $v_0$ and add $v_0$ to $V_i$.
Vertex $v_0$ represents the downsampling layer before $G_i$.
An edge $e(v_0, v)$ is constructed for each $v \in V_i$. These edges provide all possible input connections from the downsampling layer $v_0$ to all vertices $v \in V_i$.
The graph after preprocessing is denoted by $G'_i(V'_i, E'_i)$.
The edge weights in $E'_i$ are set to 1.
An example of $G'_i$ is shown in Fig.~\ref{figure:subgraph example} (b), where the new green vertex with index 0 is a pseudo vertex, representing the downsampling layer before $G_i$ in Fig.~\ref{figure:subgraph example} (a). 

After preprocessing, we enter the iterations to optimize our model by its performance (line~3 of Algorithm~\ref{alg:weight_update}).
In each iteration, we apply the sampling function in Algorithm~\ref{alg:sample_dag} on each $G'_i$ to sample the corresponding $S_i$ (line~4-5).
In the remaining paragraph, we will cover the Algorithm~\ref{alg:sample_dag}, which takes $G_i$ as the input and outputs a sampled-DAG $S_i(V_{S_i}, E_{S_i})$.
First, we initialize $V_{S_i}$ with $v_0$ (lines~1), the vertex that represents the downsampling layer.
For each vertex $v_j \in V'_i$, if $v_j$ is in $V_{S_i}$, we iterate through all its edges $e(v_j, v_k)$ to perform the edge selection (lines~3-5).
During the edge selection, for each $e(v_j, v_k)$, its edge selection probability $p$ is set to the normalized weight of $w_{e(v_j, v_k)}$ over the weights of all outgoing edges of $v_j$. 
This normalization is performed with a softmax function (line~6).
Note that a larger weight edge means a higher probability to be sampled, and the softmax function can further enhance the difference between weights and reflect it on the probability. 
If $e(v_j, v_k)$ is selected, we add $e(v_j, v_k)$ and $v_k$ into $E_{S_i}$ and $V_{S_i}$, respectively (lines~8-10).
In later iterations, the outgoing edges of $v_k$ will be extracted and performed sampling.
Finally, $S_i$ is returned after iterating through all the vertices. 
Fig.~\ref{figure:subgraph example} (b) and (c) demonstrate an example of subgraph sampling.
In Fig.~\ref{figure:subgraph example} (b), red edges in $G_1$ are selected by Algorithm~\ref{alg:sample_dag}.
According to the red edges, $S_1$ is constructed in Fig.~\ref{figure:subgraph example} (c).
Vertices without any outgoing edge are connected to the right downsampling layer.

\begin{algorithm}[!t]
 \caption{\textit{Sampling($G'_i)$}}
 \label{alg:sample_dag}
\begin{algorithmic}[1]
\Require $G'_i(V'_i, E'_i)$
\Ensure $S_i(V_{S_i}, E_{S_i})$
\State $V_{S_i} = \{v_0 \}, v_0 \in V'_i$
\State $E_{S_i} = \emptyset$
\For{each $v_j \in V'_i$}
    \If{$v_j \in V_{S_i}$}
        \For{each $e(v_j, u_k) \in E'_i$}
            \State $ p = \frac{\exp(w_{e(v_j, v_k)})}{\sum_{l=j}^7 \exp( w_{e(v_j, v_l)} ) }$
            \State random $r(0, 1)$
            \If{$p > r$} \Comment{sample by probability $p$}
                \State $V_{S_i} = V_{S_i} \cup \{v_k\}$
                \State $E_{S_i} = E_{S_i} \cup \{e(v_j, v_k)\}$
            \EndIf
        \EndFor
    \EndIf
\EndFor
\State \Return $S_i(V_{S_i}, E_{S_i})$
\end{algorithmic} 
\end{algorithm}

After sampling each $S_i$, we construct our model through the architecture in Fig.~\ref{figure:nas ea flow} and measure our evaluation metrics $\eta$ with our evaluation strategy in Algorithm~\ref{alg:weight_update} (lines~6-7).
According to $\eta$, we iterate all edges in each $G_i$ and update the edge weights which are sampled in $S_i$ in this iteration (lines~8-13).
The weight $w_e$ of edge $e$ is defined as
$$
w_e = w_e * \exp{ (\alpha (\eta - \beta)) },
$$
where $\alpha$ is the updating rate, and $\beta$ is the baseline metrics.
The weights are updated according to the difference between the evaluation metrics $\eta$ and the baseline metrics $\beta$.
We utilize an exponential function to boost the weight update.
If the sampled model has a higher performance than the baseline, the edge weights will increase by this updating equation. 
The sample probabilities of these high-performance edges will grow accordingly in the next iteration.
The baseline metrics $\beta$ will be set to the mean of the top five evaluation metrics of all history sampled models (line~14).
This setting can prompt the search process to seek higher-performance models through iterations.
This sample and weight updating process continues until the model performance converges.

\subsection{Feature Extraction} \label{sec:feature extraction}

We extract input features that correlate with routability. 
The features capture both locations of cells and the connectivity between instances, denoted as \emph{cell density features} and \emph{wire density features}, respectively.  

The cell density features include five features:
    \begin{itemize}
        \item Macro density
        \item Cell density of all cells
        \item Cell density of D Flip-Flop cells
        \item Cell density of clock tree cells
        \item Pin density in all cells
    \end{itemize}
The macro density captures the region occupied by macros. 
In addition to the cell and pin density of all cells, we note that cells with different functions would have different impacts on routability. 
To capture this effect, density distributions of the D Flip-Flips and clock tree buffers are generated separately.
These four cell density features form a 5-channel tensor in $\mathbb{R}^{w\times h\times 5}$.


For the wire density features, to differentiate nets in different sizes, nets are categorized into two groups depending on whether the fanout size is larger than a threshold. 
Then, wire density features are measured separately for each group. 
Fig.~\ref{figure:wire density} illustrates an example of each wire density feature, and the detailed descriptions are provided below:


\begin{figure} [t]
    \centering
\includegraphics[width = 0.95\linewidth]{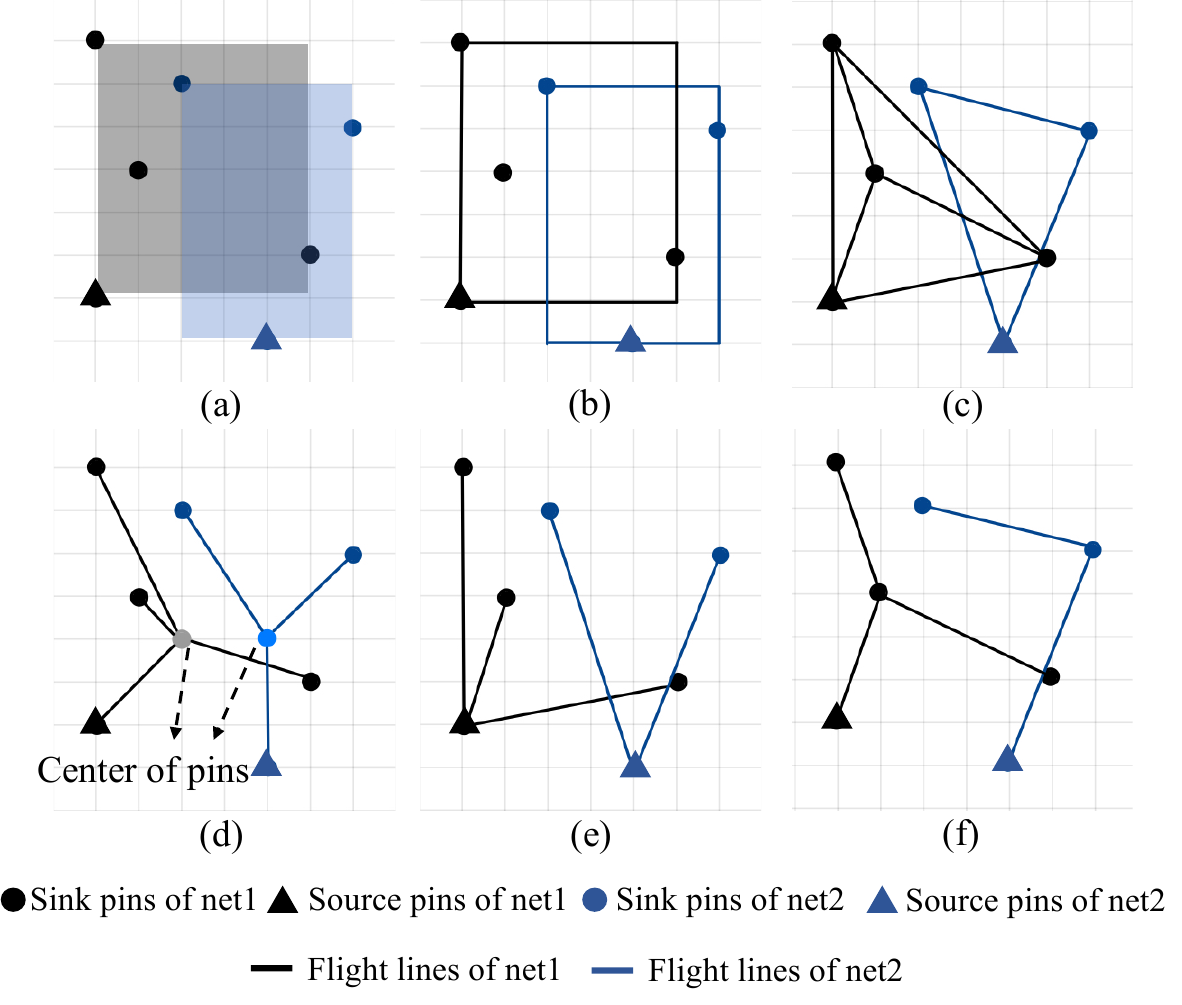}
\caption{
Wire density feature examples. (a) RUDY. (b) Bounding box. (c) Pair-wise flight lines. (d) Star flight lines. (e) Source-sink flight lines. (f) MST flight lines.
}
\label{figure:wire density}
\end{figure}

\begin{itemize}
    \item RUDY: RUDY~\cite{xie2018routenet} is derived by the total uniform wire volume spreading in the bounding box of nets in one group.

    \item Bounding box: Compared with RUDY, we directly draw the outline of the bounding box for each net in this feature. 
    We then summarize the total number of outlines passing every tile to construct the overall density map.
    
    \end{itemize}
    
    To estimate the wire density and connectivity, we adopt a heuristic named flight line, which is a line that connects two pins. Flight lines are used as features in \cite{YuDAC19}, but only for routability prediction on FPGA. In comparison, we adopted four different types of flight lines that reflect wire congestions in ASIC designs. The density of each flight line type forms a two-dimensional feature map $\in \mathbb{R}^{w\times h}$.
    \begin{itemize}
        \item Pair-wise flight lines: 
        For each net, the pair-wise flight lines are connected from each pin to all other pins of the same net.

        \item Star flight lines: For each net, the star flight line connects each pin to the center of the pins in the net.
        
        \item Source-sink flight lines: 
        In timing optimization, tools tend to connect sinks with the source through the shortest path.
        To capture this effect, source-sink flight line connects the source pin with all sink pins in the same net.
        
        \item MST flight lines: The three aforementioned flight line features tend to overestimate the routing usage. In traditional routing, minimum spanning tree (MST) is an effective algorithm to guide the router. Thus, the flight line connects all the edges in its MST.
    \end{itemize}
    
With 6 different types of features and 2 groups of the nets, there are 12 two-dimensional feature maps for wire density features. In summary, all above features are stacked together to form an input feature tensor $X \in \mathbb{R}^{w\times h\times 17}$.


\begin{table*}[!t]
  \footnotesize
  \centering
   \renewcommand{\arraystretch}{1.1}
  \caption{Comparison of the violated net count prediction}
  \label{tab: violated net cross-validation}
   \resizebox{0.98\linewidth}{!}{%
  \begin{tabular}{| c || c | c | c | c || c | c | c | c | c | c || c | c | }
      \hline
      \multirow{2}{*} {Models} & \multicolumn{4}{c||}{Kendall's $\tau$ on designs (\#nets)} & Kendall's $\tau$ & Pearson's correlation \\
      \cline{2-5}
    & s349 (270) &  mem\_ctrl (9.3k) & b17 (33.8k) & DSP (73.1k)   &  on all 74 designs  &   on all 74 designs \\     
      
      \hline \hline
	  RouteNet~\cite{xie2018routenet} & 0.3620 & 0.1547 & 0.1779 & 0.4414  & 0.5264 &   0.7224\\
	  \hline
	  \textbf{NAS-crafted model} & \textbf{0.6369} &  \textbf{0.4657} & \textbf{0.2683} & \textbf{0.7302}  & \textbf{0.5572} & \textbf{0.7930} \\
	  \hline
  \end{tabular}
   }
  \vspace{-2pt}
\end{table*}

\begin{table*}[!b]
  \footnotesize
  \centering
   \renewcommand{\arraystretch}{1.1}
  \caption{Comparison of the DRC hotspot detection}
  \label{tab: drc hotspot cross-validation}
   \resizebox{0.85\linewidth}{!}{%
  \begin{tabular}{| c || c | c | c | c || c | c | c | c | c | c || c | }
      \hline
      \multirow{2}{*} {Models}   & \multicolumn{4}{c||}{ROC-AUC on designs (\#nets)} & \multirow{2}{*} {ROC-AUC on all 74 designs} \\
      \cline{2-5}
        & s349 (270) &  mem\_ctrl (9.3k) & b17 (33.8k) & DSP (73.1k) & \\
      \hline \hline
	  RouteNet~\cite{xie2018routenet} & 0.829 & 0.844 & 0.902 & 0.866  & 0.847 \\
	  \hline
	  PROS~\cite{chen2020pros} & 0.487 & 0.483 & 0.478 & 0.489 & 0.676 \\
	  \hline
	  cGAN~\cite{YuDAC19}  & 0.516 & 0.515 & 0.521 & 0.517 & 0.510 \\
	  \hline
	  \textbf{NAS-crafted model}    & \textbf{0.865} & \textbf{0.891} & \textbf{0.911} & \textbf{0.884} & \textbf{0.865} \\
	  \hline
  \end{tabular}
   }
  \vspace{-6pt}
\end{table*}

\section{Experimental Results}
In this section, we first describe our experimental setups on dataset construction, neural architecture search, and feature extraction along with training details. We then present our evaluations on two routability prediction benchmarks: violated net count and DRC hotspot detection.

\subsection{Experiment Setup}



\noindent \textbf{Dataset Construction.} 

We construct a comprehensive dataset using 74 designs with largely varying sizes from multiple benchmarks. There are 29 designs from ISCAS’89~\cite{brglez1989combinational}, 13 designs from ITC'99~\cite{corno2000rt}, 19 other designs from Faraday and OpenCores in the IWLS'05~\cite{albrecht2005iwls}, and 13 designs from ISPD'15~\cite{bustany2015ispd}. 
For each design, multiple placement solutions are generated with different logic synthesis or physical design settings. Altogether 7,000 placement solutions are generated from these 74 designs. 
We apply Design Compiler\textsuperscript{\textregistered} for logic synthesis and Innovus\textsuperscript{\textregistered}~\cite{Innovus} for physical design with the NanGate 45nm technology library~\cite{URL:NanGate}. 
The input feature maps are collected at the post-placement stage, and the ground-truth DRC results are available after detailed routing finishes.



\noindent \textbf{NAS and Accuracy Measurement.} 

We adopt the NAS in Section~\ref{sec:NAS} to explore the search space defined for the violated net count prediction and the DRC hotspot detection. 
The constructed dataset is firstly separated into two splits of different designs. 
The training split contains layouts from 51 designs, and the validation split contains layouts from 23 designs. 
This is the setting for the evaluation strategy during our NAS process.
Then the final model evaluation is performed under 5-fold cross-validation. 
More specifically, our dataset is randomly separated into 5 folds of different designs. 
We perform validation for each fold after training our model with all other 4 folds. 
Finally, we average the validation results for each fold and obtain the 5-fold cross-validation metrics. 
Note that the placement layouts generated from the same designs are assigned to the same fold for the fairness of evaluation.
With cross-validation, we can thoroughly measure the generalization capability of our model in all partitions of the dataset. 
To maximize the performance of our NAS-crafted models, we select the top-5 models produced by our NAS method and pick up the top-performance one with best cross-validation criteria as the final NAS-crafted model.

The overall search process runs for 0.3 days on 8 NVIDIA TITAN RTX GPUs with Intel\textsuperscript{\textregistered} Xeon\textsuperscript{\textregistered} E5-2687W CPU. Note that an ML expert may take up to 2 months to design a promising CNN model based on~\cite{xie2018routenet}.
This search time shows that NAS process substantially shortens the development cycle of ML models on EDA prediction.

\noindent \textbf{Feature Extraction \& Training.} 

For each layout in the target dataset, we follow Section~\ref{sec:feature extraction} to perform feature extraction. Each feature tensor uses $224\times 224$ resolution and contains 17 channels.
We employ the following hyperparameters to conduct model training for both searching and final evaluation in our experiments:
We train our model for 45 epochs with Adam optimizer~\cite{kingma2014adam}, a batch size of 48, and a fixed learning rate of $0.0005$. 
To combat overfitting and improve generalization, we use a L2 weight decay of $10^{-5}$ and ReLU activation.

\noindent \textbf{Baseline Methods.} 

We implement multiple representative or state-of-the-art routability estimators by our own and use them as baselines to compare with our solutions. 
They include RouteNet~\cite{xie2018routenet}, PROS~\cite{chen2020pros}, and the cGAN-based method~\cite{YuDAC19}. 
We compare the DRC hotspot detection accuracy with all three baselines. 
In comparison, only RouteNet~\cite{xie2018routenet} proposes its solution on estimating the routability of the whole placement. Thus, for violated net count prediction, we can only measure and compare with the accuracy of RouteNet~\cite{xie2018routenet}. 


\subsection{Violated Net Count Prediction} \label{sec:violated net evaluation}
In this section, we evaluate the effectiveness of our model on violated net count prediction using the metrics Kendall's $\tau$ in the 5-fold cross-validation.
For the baseline method, RouteNet~\cite{xie2018routenet} utilizes ResNet-18 structure to predict the overall routability. 
In practice, during chip design, designers typically focus on optimizing a specific design and care about the routability between different layout solutions of the same design. 
Therefore, the prediction performance within each design is important. 
In TABLE~\ref{tab: violated net cross-validation}, we first compare Kendall's $\tau$ evaluated on the placement solutions of some specific designs.
We select four designs with distinct number of nets ranging from 270 to 73.1k in the target dataset and show their Kendall's $\tau$ for both RouteNet and our NAS-crafted model.
More importantly, we then report the averaged Kendall's $\tau$ over all 74 designs under 5-fold cross-validation to demonstrate the overall performance. 
Our NAS-crafted model achieves average Kendall's $\tau$ of $0.5572$ and clearly outperforms RouteNet by around $0.03$ in absolute value, which is $5.85\%$ improvement.
In addition, we also apply Pearson's correlation coefficient, another widely adopted metric for correlation measurement, to evaluate the performance. 
Our NAS-crafted model achieves average Pearson's correlation of $0.7930$, which also outperforms RouteNet by around $0.07$ in absolute value, showing $9.7\%$ improvement.

\begin{figure} [t]
    \centering
\includegraphics[width = 0.8\linewidth]{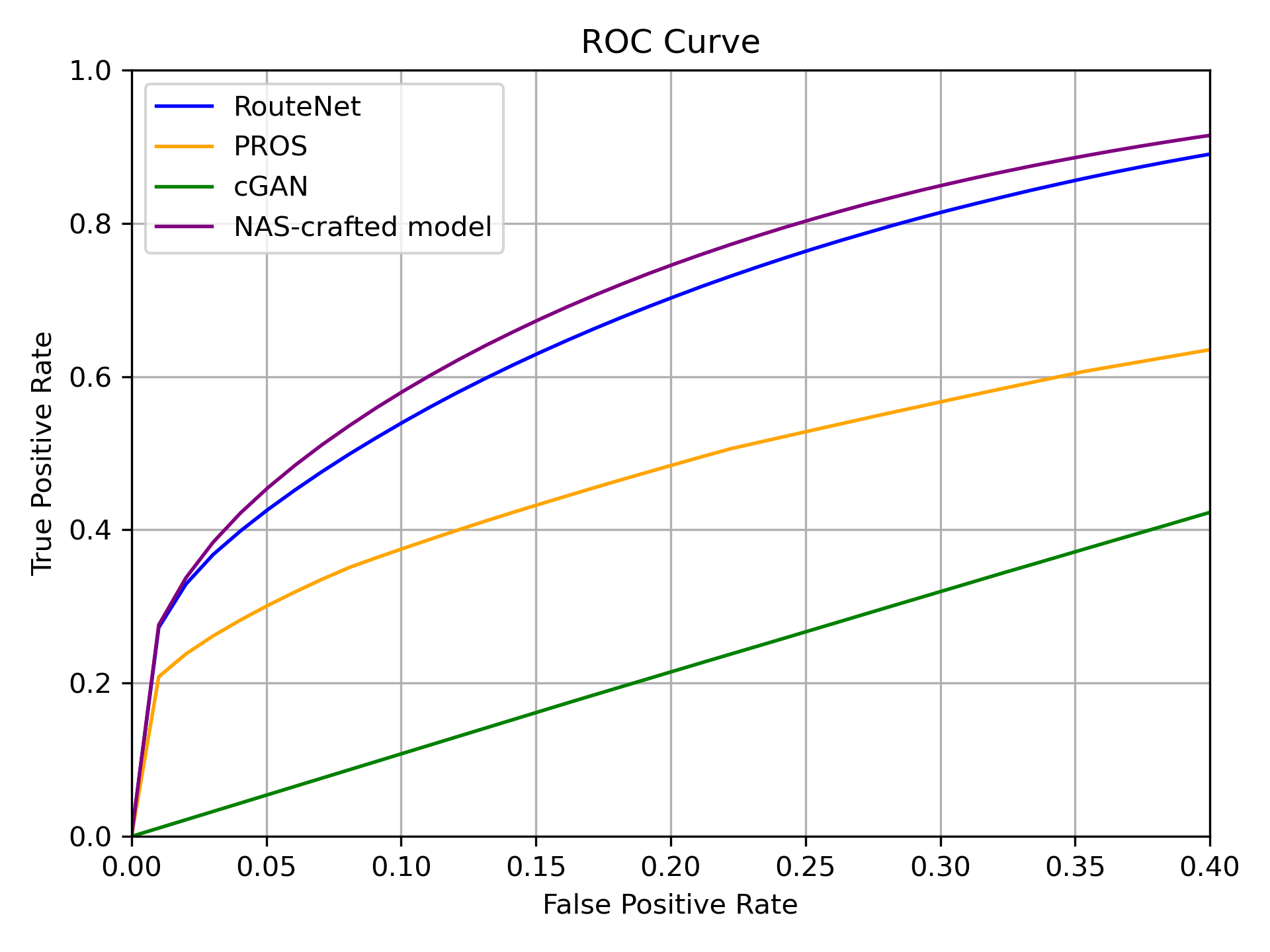}
\caption{ROC curves (focus on left-half region) measured on all 74 designs. Comparisons between the previous works RouteNet, PROS, and our NAS-crafted model.}
\label{figure:roc}
\end{figure}

\subsection{DRC Hotspot Detection}


The quality of DRC hotspot detection is measured with the ROC-AUC. 
TABLE~\ref{tab: drc hotspot cross-validation} shows the average ROC-AUC comparison with all three previous works under 5-fold cross-validation.
Our NAS-crafted model achieves ROC-AUC of $0.865$ when evaluated on all 74 designs. 
It clearly outperforms the best human-crafted model among our three baselines by $0.018$ in absolute value, which is a $2.1\%$ improvement. 
It shall be noted that this whole automatic search process takes only $0.3$ days, demonstrating the high efficiency of our search algorithm. 
Fig.~\ref{figure:roc} further demonstrates the average ROC curves, corresponding to the averaged ROC-AUC in TABLE~\ref{tab: drc hotspot cross-validation}. It focuses on the left-half region since it demonstrates most accuracy information in this case.
At the same cost of false alarms in the x-axis value, the true positive rate of NAS-crafted model at the y-axis is obviously higher than PROS and RouteNet.
TABLE~\ref{tab: drc hotspot cross-validation} also shows the ROC-AUC on the placement solutions of four specific designs (same as in TABLE~\ref{tab: violated net cross-validation}).
In each design, the NAS-crafted model reports $1.0\%$ to $4.3\%$ higher ROC-AUC than the best human-crafted model among baselines.

Previous works PROS and the cGAN-based model show significantly low ROC-AUC on many designs.
After putting our best effort in optimization and hyperparameter tuning, these are the best results we can achieve currently.
One possible reason is that compared with other baselines like RouteNet, PROS and cGAN, architectures are over-complex, and thus their models may suffer from the data heterogeneity among training and testing designs from different benchmarks.

\begin{figure*} [!t]
\centering
\includegraphics[width = 0.9\textwidth]{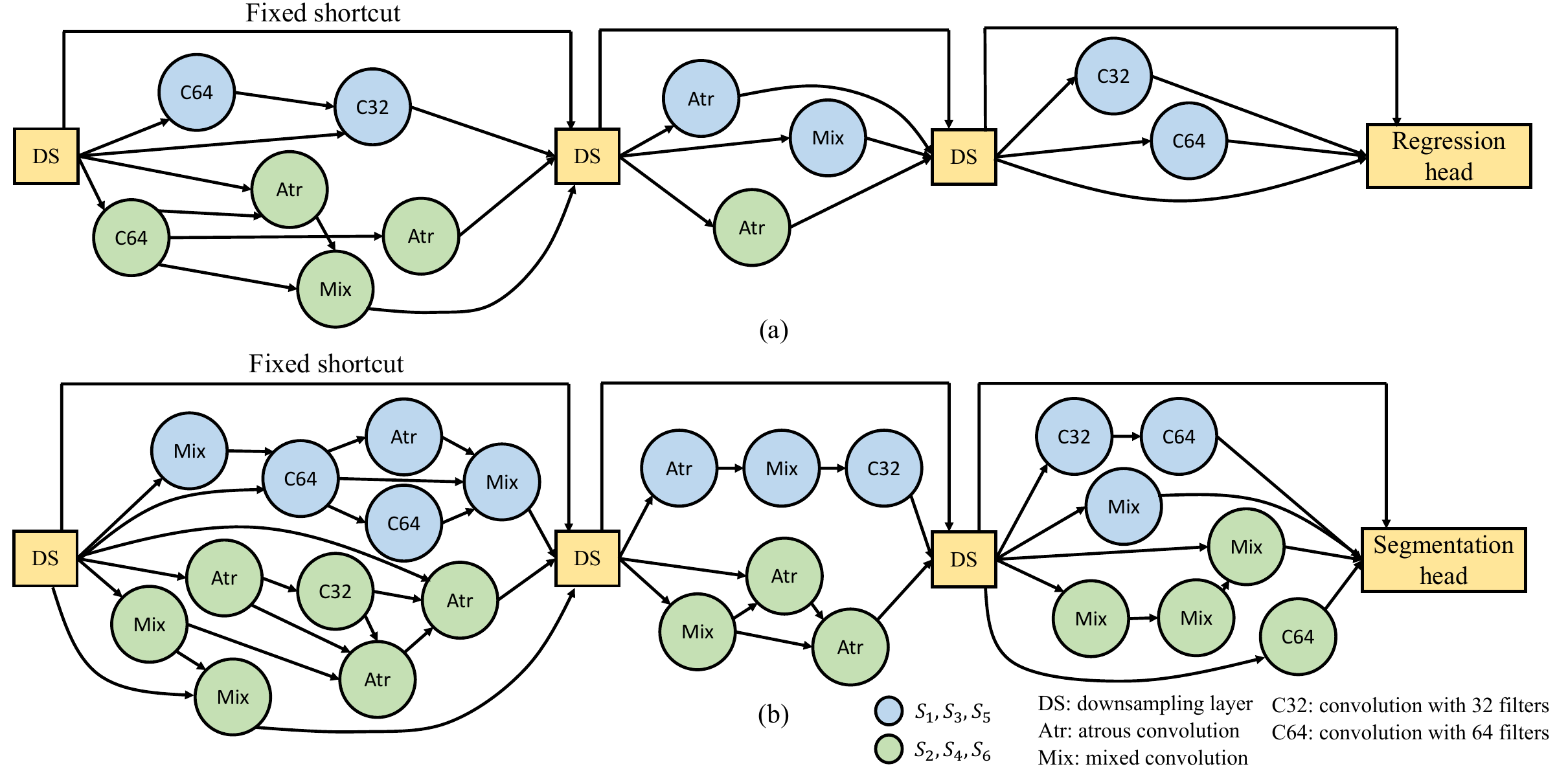}
\caption{
(a) The NAS-crafted model for violated net count prediction. (b) The NAS-crafted model for DRC hotspot detection. 
}
\vspace{-1em}
\label{figure:NAS model}
\end{figure*}

\section{Discussion}

We compare the structures of our NAS-crafted models and the manually-crafted models (e.g., RouteNet~\cite{xie2018routenet}, PROS~\cite{chen2020pros} and cGAN~\cite{YuDAC19}), and give a brief analysis. This may provide some insights to future routability estimator development. 

\vspace{0.2em} \noindent \textbf{Violated Net Count Prediction.}

Fig.~\ref{figure:NAS model}(a) shows the most promising architecture discovered by our NAS method for violated net count prediction.
As mentioned in Fig.~\ref{figure:nas ea flow}, there is a fixed shortcut from each DS layer to the next DS layer or regression/segmentation head.
In sampled-DAGs 1 and 2, we observe that using wider convolution layers (larger number of filters) as parent vertices can help to improve performance because later operations can exploit richer feature representation.
In sampled-DAGs 3 and 4, due to the nature of violated net count prediction, our NAS method prefers a compact structure composed of only two atrous convolutions and one mixed convolution.
Since both atrous convolution and mixed convolution can utilize large-scale input patterns by increasing the receptive field, they are good fits of distilling the features that spans a wide layout region.
Finally, in sampled-DAGs 5 and 6, the two regular convolutions with different numbers of filters and the shortcuts simply enrich the representation extracted by the previous operations and pass it to the regression head.

However, human developers can hardly explore structures that are similar to the NAS-crafted model.
For violated net count prediction, most human-crafted models only support a limited number of operators (typically regular convolution), and thus have limited ability to learn the large-scale input patterns.
They also adopt a highly hierarchical architecture that lack the ability to aggregate different levels of features.
In contrast, our NAS method supports operators that process features very differently.
The variation of operations of vertices on the branches greatly increase the diversity of feature representations that can be explored by our NAS method.
Moreover, it can also construct a number of scalable parallel branches and explore flexible interactions among them, which is inherent in the topology of the guide-DAGs.
Thereby, our NAS method can extract feature representations of large-scale patterns, which is critical to improving the accuracy of violated net count prediction.


\vspace{0.2em} \noindent \textbf{DRC Hotspot Detection.}

Fig.~\ref{figure:NAS model} (b) shows the most promising architecture discovered by our NAS method for DRC hotspot detection.
In sampled-DAGs 1 and 2, we observe that our NAS method adopts a complex model with rich interactions among atrous convolution, mixed convolution, and regular convolution with different filters.
Such structure is able to extract both small-scale and large-scale input patterns by utilizing rich interactions among parallel branches.
In sampled-DAGs 3 and 4, the structure is much more compact than in Sampled-DAGs 1 and 2, but all branches contain at least one atrous convolution layer, which highlights the importance of larger receptive fields of the filters.
In sampled-DAGs 5 and 6, we observe that the NAS highly prefers a complex combination of mixed convolution layers, which effectively learns from both small-scale and large-scale input patterns. 
Such structure reflects the nature of DRC hotspot detection that both local patterns and global patterns of a layout influences the routability at each point. Similar to violated net count prediction, human-crafted models~\cite{xie2018routenet, chen2020pros, YuDAC19} designed for DRC hotspot detection also adopt highly hierarchical architectures with limited types of operations and thus have similar disadvantages.

Compared with violated net count prediction, DRC hotspot detection is much more challenging because it requires detailed pixel-wise prediction of hotspot locations. Such an essential difference between these two routability prediction tasks is clearly captured and reflected by the two discovered models in Fig.~\ref{figure:NAS model} (a) and (b). 
Our search algorithm generates a significantly more complex model for DRC hotspot detection in Fig.~\ref{figure:NAS model} (b), especially near its output at segmentation head. 
These complex structures help to better utilize input patterns of various scales and thus improve the accuracy of the fine-grained prediction on DRC hotspot locations.

\section{Conclusion}

In this paper, we propose a NAS-based method to automate the design of ML models for two routability prediction applications. We believe this is the first research effort on automatic ML development for EDA problems. Based on a large search space with various operations and highly flexible connections, the NAS method efficiently generates a high-performance model in 0.3 days. The automatically generated model proves to outperform previous representative routability estimators on a large dataset.

Besides proposing the NAS-based method, we provide our analysis on the automatically generated model and hope it benefits future routability estimator development. In addition, considering the similarities in solutions between routability estimation with other essential ML for EDA problems like IR drop estimation, clock tree prediction, lithography hotspot detection, optical proximity correction, etc., NAS-based solutions may ultimately also benefit the solving of these problems.

\section*{Acknowledgments}
This work is supported by Cadence Design Systems Inc. through the NSF IUCRC 1822085 for Alternative Sustainable and Intelligent Computing (ASIC), NSF 2106828, and
Semiconductor Research Corporation Tasks 2810.021 and 2810.022 through UT Dallas’ Texas Analog Center of Excellence (TxACE).

\bibliographystyle{IEEEtran}
\bibliography{references}
\end{document}